# Diachronic Variation in Grammatical Relations

*Aaron Gerow    Khurshid Ahmad*
Trinity College Dublin, College Green, Dublin, Ireland.
`gerowa@tcd.ie, kahmad@scss.tcd.ie`

ABSTRACT
We present a method of finding and analyzing shifts in grammatical relations found in diachronic corpora. Inspired by the econometric technique of measuring return and volatility instead of relative frequencies, we propose them as a way to better characterize changes in grammatical patterns like nominalization, modification and comparison. To exemplify the use of these techniques, we examine a corpus of NIPS papers and report trends which manifest at the token, part-of-speech and grammatical levels. Building up from frequency observations to a second-order analysis, we show that shifts in frequencies overlook deeper trends in language, even when part-of-speech information is included. Examining token, POS and grammatical levels of variation enables a summary view of diachronic text as a whole. We conclude with a discussion about how these methods can inform intuitions about specialist domains as well as changes in language use as a whole.

# 1 Introduction

Language is both representative and constitutive of the world around us, which makes tracking changes in its use a central goal in understanding how people make sense of the world. Charting these changes is a two-part challenge: extracting meaningful, diachronic data and finding the best way to characterize it. Literature on visualizing themes in text (Havre et al., 2002), identifying topics (Kim & Sudderth, 2011; Rosen-Zvi et al., 2010) and analyzing the sentiment of financial news and social media (Tetlock, 2006; Kouloumpis et al., 2011) are examples of how changes in language are linked to changes in the world. The underlying assumption is that shifts in the distribution of words and phrases may indicate changes in a domain or community.

Language use in specific subjects is known to be productive: a relatively small set of words are not used repetitively, instead, they give rise to new words through inflectional and derivational processes (Halliday & Martin, 1993). This productivity, as genesis and obsolescence, suggests that by analyzing diachronic text we can gain insight into the ontological commitment of a domain (Ahmad, 2000; see McMahon, 1994 for general language and Geeraerts, 2002 for scientific language). Topic modeling has been shown to make use of frequency observations to build probabilistic models with which to infer clusters of representative words (Griffiths & Steyvers, 2004). However, word-frequency is only one level of linguistic variation. Other shifts, like part-of-speech and grammatical relations, are also important in understanding a domain's language. As we will see, some trends in frequency have consistent underlying trends in grammatical relations that signal changes not apparent at higher levels.

By organizing text diachronically, frequency data can be analyzed as a time-series. Enabled by an endless amount of text on the internet, corpus linguists have constructed large databases of such text to chart linguistic trends (for example Davies, 2010). Sentiment and opinion mining have developed nearly real-time methods of tracking sentiment in text (Tetlock, 2007). Other work has tracked shifts in parts-of-speech (Mair et al., 2003) and related fluctuations in verb-distributions to stock-markets (Gerow & Keane, 2011). Perhaps the boldest claim analysts of language-change have made, is that by analyzing the relative frequency of words over time, we gain a quantitative view of culture itself (Michel et al., 2010).

To find variation over time, we explore whether a time-series analysis can help uncover patterns of seemingly random movements in frequency. To do this, we use continuously compounded return and volatility. These measures are commonly used in econometrics where high prices tend to beget higher prices and low prices, lower still. This phenomenon of auto-correlation is also apparent in frequency-variations in text, which means an analysis of mean and variance can be misleading. Using return and volatility has been used in sentiment analysis, where it was found that negative-affect terms caused a larger, and longer-lasting deviation from the mean than positive terms (Ahmad, 2011). To our knowledge, these metrics have not been used to investigate trends in words with respect to the grammatical relations in which they are found. By looking at grammatical relations in particular, we get a picture *how* those words are used. This type of analysis may shed light on language-change and perhaps help predict trends in topics and key-terms which characterize a domain.

The driving question in this paper is whether second-order analyses of diachronic text can be used to find trends not apparent on the surface. Using return and volatility we get a synoptic picture of changes in a diachronic corpus which is informed by the *kind* of changes themselves. And by examining grammatical relations, we note specific shifts not apparent at the lexical token level. Our results offer some interesting findings about academic language: by analyzing key terms, we find discernible trends at varying levels of language as well as generalizations about the text as a whole.

## 2   Methods: Measuring Diachronic Shifts

A series, $f(t)$, is a discrete set of ordered data-points which typically exhibit a degree of auto-correlation, meaning that preceding values tend to have a discernible effect on subsequent values. Even in heteroskedastic series – the log-normal regression of which is non-linear – the values of the past, $f(t-n)$, tend to be good predictors of successive values. Measuring and making use of this relation is the focus of predictive econometric models, widely employed in economics and finance (Taylor, 2005). One common method used to measure the variation in a time-series is to calculate successive ratios of consecutive values, known as the *return* of a series. Unlike standard deviation in a sample or population, calculating the return series is order-aware and can be computed for varying segments of time and degrees of resolution. In our analysis we use the continuously compounded return defined as:

$$r(t) = \log \frac{f(t)}{f(t-1)} \quad (1)$$

Unlike the original series, returns are not serially correlated. This leads economists to consider variance in the return-series, or *volatility*, a better way to estimate the dispersion of values in the original. For a time-series, $f(t)$ of $N$ ordered-points, volatility is defined as:

$$v = \sum_{t=0}^{N} \frac{(r(t) - \bar{t})^2}{N(N-1)} \quad (2)$$

We can combine equations 1 and 2 to gain a view of the overall variation in a corpus composed of an time-ordered set of documents, $D$, as the mean of continuously compounded returns, $\bar{r}$:

$$\bar{r} = \frac{1}{|D|} \sum_{d \in D} \log \frac{f_d}{f_{d-1}} \quad (3)$$

where $f_d$ is a frequency observation of document $d$.

## 3   Results: Variation in the NIPS corpus

The NIPS corpus[1] consists of papers from thirteen volumes of Neural Information Processing Systems proceedings. It contains 6.7 million words in 1,740 documents published over 13 years from 1987 to 1999, with an average of 516,394 tokens per year. The mean return for yearly corpus-size was 4% with a volatility of 11% – exhibiting a relatively slow, steady growth.

---

[1] Available at http://www.cs.nyu.edu/˜roweis/data.html.

Sketch Engine (Kilgarriff et al., 2004) was used to clean, lemmatize, tag, divide the corpus into yearly sub-corpora and provide frequencies of grammatical relations (see Table 1). Sketch Engine uses a pre-trained version of TreeTagger (Shmid, 1994) and was also used to extract the grammatical relations by applying abstract tag templates or a "sketch grammar" to the tagged corpus. It should be noted that the NIPS corpus offers a uniformly diachronic corpus on which to test our methods, but any diachronically organized corpus would suffice. Moreover, though we rely on a POS tagger and subsequent POS-based grammatical relation extraction, the method, as such, is applicable to any language from which grammatical relationships can be extracted.

By comparing relative frequencies to the ACL Anthology Reference Corpus (ACL ARC)[2], we isolated five terms to analyze in detail: *network*, *learning*, *training*, *algorithm* and *neuron*. The relative frequency and return series are shown in Figure 1 and their respective mean frequency, standard deviation, mean of returns and volatility are given in Table 2. Taken together, we have an overview of how the use of these terms changed over the thirteen-year corpus. We can see that *algorithm* doubled in usage, while *neuron* showed a steady decline and *network* a turbulent decline from dominating the five words. Only *learning* was steady throughout. Also note how *network* dominates the plot of relative frequency, despite having a relatively steady return series. Alternatively, *algorithm* appears quiescent in the frequency series, but shows considerable fluctuations in return.

| Relationship | Example |
|---|---|
| `subject_of` | Cooperative **training** *gives* a framework [...] |
| `object_of` | *Showing* that **training** increased the [...] |
| `modifier` | [...] with a single **training** *pattern*. |
| `a_modified` | [...] at *smaller* **training** set sizes. |
| `n_modified` | [...] using back *propagation* **training**. |
| `and/or` | [...] achieved during **training** *or* testing. |

Table 1: Common grammatical relations found in our analysis. The defining feature of each is italicized in the example and the word-in-question is in bold.

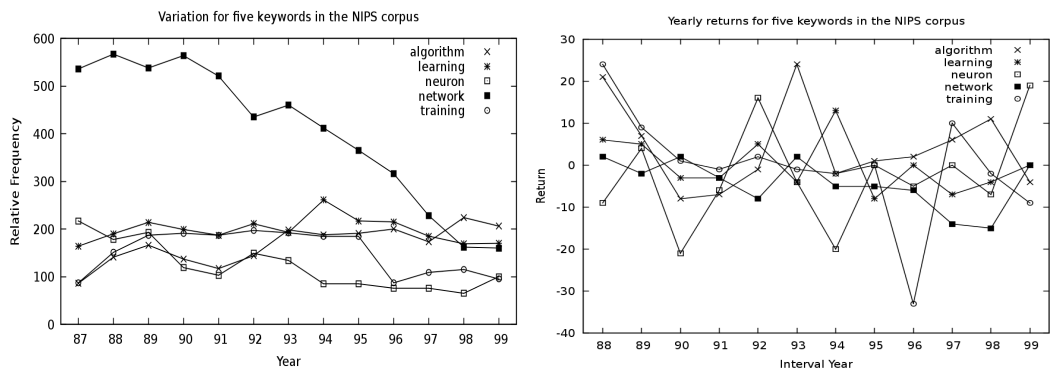

Figure 1: On the left are the relative frequencies (per 100,000 tokens) for five keywords (all forms) in the NIPS corpus. On the right are the return series for each keyword. Note how there is considerably more variance in the return series than in the relative frequencies – particularly for *neuron* and *training*.

---

[2]Available at http://acl-arc.comp.nus.edu.sg/.

|  | $\bar{f}$ | SD($f$) | $\bar{r}$ | $v$ |
|---:|---:|---:|---:|---:|
| *training*-[n] | 0.151% | 30% | 0.3% | 13% |
| *neuron*-[n] | 0.122% | 40% | -3% | 12% |
| *algorithm*-[n] | 0.165% | 27% | 3% | 10% |
| *learning*-[n] | 0.198% | 13% | 0.1% | 6% |
| *network*-[n] | 0.405% | 37% | -4% | 6% |

Table 2: Summary statistics for the relative frequency and return series of the five keywords we examined in the NIPS corpus, ordered by volatility. Shown are the relative frequency ($\bar{f}$; per 100,000 tokens), the standard deviation of the frequency, the mean of return ($\bar{r}$; Eq. 3) and the volatility ($v$; Eq. 2).

The key question in this paper is whether there is significant variation at the grammatical level. In Table 3, the nouns *learning* and *training* are presented with a breakdown of their occurrences and their most common grammatical relations. Note that although *learning*-[noun] appears steady (1-lag auto-correlation = 28%, $p = 0.18$), it exhibits relatively high volatility in its two most common relations: adjective modified and as a modifier. *Training*-[noun], which at 1-lag is 61% auto-correlated ($p < 0.1$), is also deceptively summarized by its frequencies being found in a number of a volatile relationships, the least volatile being the one which increased the most: as a modifier.

The remaining three terms, *network*, *algorithm* and *neuron*, are presented in Figure 2, which contains plots of the mean return against volatility for each POS-class and the five most common relations in which they occur, as well the plots of the relative frequency throughout the corpus. Consider *network* in Figure 2, which, despite showing a steady negative trend overall in Figure 1, is increasingly modified by both nouns and adjectives, appearing less frequently as a subject. Also consider forms of the word *neuron*, which include the two adjectives *neural* and *neuronal* both in stable states compared to its noun forms. Though *neuron* declined in use overall, it shows wide variation in two relations, `and/or` and `modifier`, in addition to an increase in *neuron*-[noun] being noun-modified.

| Year: | 87 | 88 | 89 | 90 | 91 | 92 | 93 | 94 | 95 | 96 | 97 | 98 | 99 | $\bar{f}$ | $\bar{r}$ | $v$ |
|---:|---:|---:|---:|---:|---:|---:|---:|---:|---:|---:|---:|---:|---:|---:|---:|---:|
| *learning*-[n] | | | | | | | | | | | | | | | | |
| N | 164 | 190 | 214 | 199 | 187 | 211 | 193 | <u>261</u> | 217 | 215 | 185 | 169 | 170 | **198** | **0.1%** | **6%** |
| a_modified | 14 | 23 | 24 | 28 | 25 | 26 | 19 | <u>42</u> | 29 | 31 | 27 | 23 | 25 | **26** | **5%** | **32%** |
| modifier | 19 | 20 | 19 | 20 | 18 | 19 | 18 | 24 | <u>27</u> | 23 | 16 | 15 | 22 | **20** | **1%** | **20%** |
| n_modified | 10 | 8 | 11 | 14 | 11 | 17 | 11 | <u>23</u> | 17 | 18 | 21 | 17 | 16 | **15** | **2%** | **15%** |
| object_of | 8 | 10 | 11 | 13 | 10 | 9 | 8 | <u>14</u> | 9 | 12 | 9 | 7 | 8 | **10** | **0%** | **12%** |
| subject_of | 7 | 6 | 8 | 14 | <u>20</u> | 18 | 12 | 19 | 15 | 16 | 13 | 9 | 10 | **13** | **1%** | **15%** |
| *training*-[n] | | | | | | | | | | | | | | | | |
| N | 87 | 151 | 187 | 191 | 187 | <u>197</u> | 192 | 185 | 185 | 87 | 109 | 115 | 95 | **151** | **0.3%** | **13%** |
| a_modified | 7 | 7 | 9 | 8 | 11 | <u>15</u> | 11 | 8 | 12 | 9 | 5 | 8 | 4 | **9** | **-1%** | **42%** |
| modifier | 34 | 57 | 71 | 88 | 103 | 95 | 91 | 95 | 164 | <u>167</u> | 111 | 96 | 74 | **96** | **7%** | **25%** |
| n_modified | 5 | 4 | 5 | 7 | 9 | 14 | 10 | 8 | <u>16</u> | 13 | 5 | 6 | 4 | **8** | **-1%** | **44%** |
| object_of | 9 | 9 | 8 | 13 | 14 | 17 | 16 | 17 | <u>33</u> | 24 | 15 | 14 | 7 | **15** | **-1%** | **33%** |
| subject_of | 10 | 10 | 14 | 15 | 18 | 15 | 22 | 25 | <u>40</u> | 25 | 15 | 12 | 9 | **18** | **-1%** | **30%** |

Table 3: Variation in the relative frequency of *training*-[noun] and *learning*-[noun] and occurrences in their five most common grammatical relationships. Here $\bar{f}$ is the mean relative frequency. Volatility ($v$) and mean return ($\bar{r}$) are calculated as in equations 2 and 3 respectively. Maximum values are underlined and summary statistics are shown in bold.

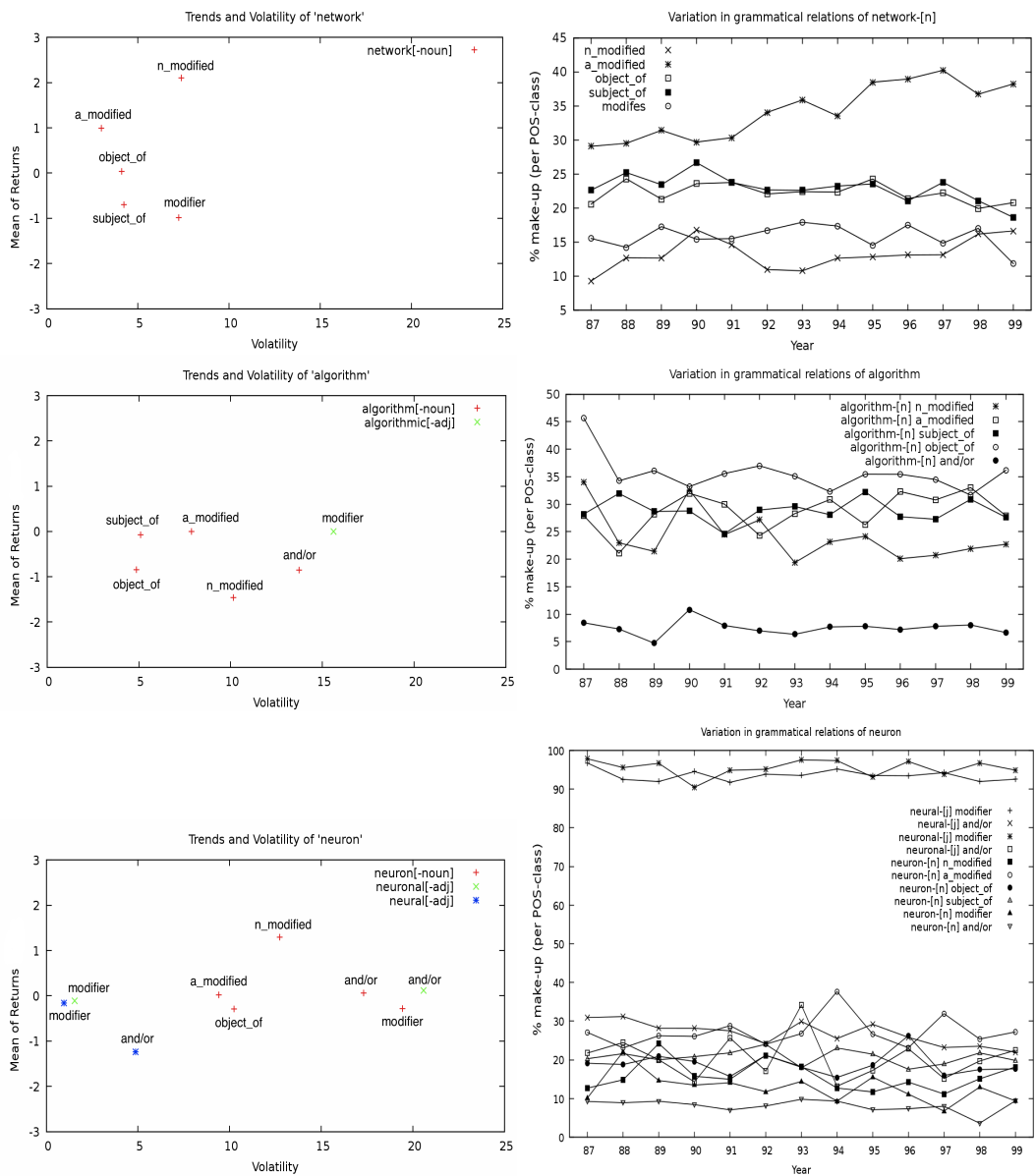

Figure 2: Grammatical shifts for the three keywords *network* (top), *algorithm* (middle) and *neuron* (bottom) are summarized in each pair of plots. On the left, the mean of returns (Eq. 3) is plotted using the word-form's relative frequency (per 100,000 tokens) against its volatility (Eq. 2) for each relationship in which it was found. In these plots, we expect clusters around the origin, where relations show little trend or volatility. On the right are plots showing the percentage make-up of each relationship over the NIPS corpus. Note that relationships which occurred less than 10 times per year are not shown but are factored into the percentage calculations.

Looking at individual words can tell us a lot about how their independent usage changes over time, but it is also useful to take a broad look at a corpus to. To this end, we extracted the top 200 noun keywords in the NIPS corpus, again by comparing their distributions to the ACL ARC. We excluded nouns that occurred less than 10 times per 100,000 tokens each year, and those that did not occur in at least two different grammatical relations every year. Because some commonly cited authors dominate the noun-distribution when compared to the reference corpus, we also removed proper nouns. In the end we were left with 43 nouns which were indicative of NIPS and consistent enough to ensure a complete analysis.

The most common relations were the same as those for *training, learning, neuron, algorithm* and *network*. For each noun we computed the mean return, volatility and the correlation between the relationship and the frequency of the overall word (all forms). For mean return, only `and/or` was significantly changing at 0.7% ($p < 0.1$). The most volatile relationship among the nouns was `n_modified` which had 27.5% volatility. However, the least volatile relationship, `subject_of`, showed 15.3% volatility. Lastly, no relationship consistently correlated with words' overall usage. In fact, the correlations themselves were highly variable (SD = 37.03%), implying that grammatical relationships are not independently indicative of word usage.

To further explore whether grammatical relationships could be indicative of a word's change in usage, we grouped words with positive trends and compared them to words with negative trends. Of the 43 nouns, 17 had a positive mean return ($\bar{r} \geq .5$), 21 had negative mean return ($\bar{r} \leq -.5$) and 5 were relatively steady ($-.5 < \bar{r} < .5$). We found that in negative trends `subject_of`, `object_of` and `a_modified` were more likely to peak *after* the word as a whole (`subject_of`: 11 proceeding the word's peak, 3 preceding; `object_of`: 13 and 6; `a_modified`: 15 and 3). The frequency of preceding and proceeding a word's peak were weighted by the number of positive to negative trends (40% positive and 49% negative). Using the weighted scores, we found that in negative trends `and/or` comparison was 7.4 times more likely to proceed the word's peak when compared to positive trends. On the other hand, in positive trends, both adjective and noun modification were 3.3 times more likely to peak before the word as a whole than in negative trends. Results of this analysis are presented in Table 4.

| Relationship | Preceded | Simultaneous | Proceeded |
|---:|---|---|---|
| *Positive Trends (N=17)* | | | |
| subject | 2.8 (7) | 0.0 (0) | 1.2 (4) |
| object | 5.5 (14) | 0.0 (0) | 2.8 (7) |
| a_modified | 4.7 (12) | 0.8 (2) | 2.0 (5) |
| n_modified | 3.2 (8) | 0.8 (2) | 2.0 (5) |
| modifier | 3.6 (9) | 0.0 (0) | 2.0 (5) |
| and/or | 4.0 (10) | 0.0 (0) | 0.8 (2) |
| *Negative Trends (N=21)* | | | |
| subject | 1.5 (3) | 0.0 (0) | 5.4 (11) |
| object | 2.9 (6) | 0.0 (0) | 6.3 (13) |
| a_modified | 1.5 (3) | 0.5 (1) | 7.3 (15) |
| n_modified | 1.0 (2) | 1.0 (2) | 4.9 (10) |
| modifier | 1.5 (3) | 0.5 (1) | 5.4 (11) |
| and/or | 2.0 (4) | 0.0 (0) | 5.9 (12) |

Table 4: Of the 43 noun keywords, shown here are the weighted frequencies of how many times a given relationship's frequency peaked before, simultaneously and after the word's overall frequency. The weighting was done by the number of trends in each category (17 positive and 21 negative). Raw frequencies are shown in parentheses.

## 4 Analysis & Discussion

One example of how shifts in relations indicate changes in the domain is the increased noun modification of *neuron*. Since the beginning of the NIPS corpus in 1987, a great deal of research has been undertaken to discern and simulate the functions of various neurons in the brain. The increased noun modification of *neuron* may be due to increased attention to particular types and functions of neurons. The word *network* also exhibits this change to being increasingly noun-modified but has steadily decreased in use as a subject. This could be due to the ubiquity of the term in the NIPS community; no longer is a neural network a "network" as such, but something more specific, like a "self organizing map", a "connectionist model" or "multilayer perceptron." Lastly, recall that despite the overall steadiness of the word *training* (Table 2), its use as a modifier dominates its ascent. This could be because the concept of training became established midway through the corpus, enabling terms like "training sample" or "training data" without as much explanation of training specifically. Though these results are somewhat speculative in nature, we feel they go deeper than first-order analyses of frequencies, by measuring the changes through the corpus as a whole.

The broader analysis of 43 key-nouns exemplifies some techniques for uncovering how changes at different levels of language use may be interrelated. We did not find a grammatical relationship among the key nouns that consistently correlated with the term's use, which implies that grammatical variation is informed by the lexicon. Comparing rising and falling patterns, we found that words which are increasingly common tend to be preceded by increased modification, both adjectival and nominal. Perhaps this points to the need for authors to further specify concepts before the community adopts them. Conversely, terms which were decreasing in use were more likely to see a subsequent peak in `and/or` comparison. This may point to an explanatory transition from one term to another, that is, writers liken new terms to old terms fading from use.

The key observation in this paper is that academic language – which is used primarily used to explain complex, technical ideas – exhibits grammatical shifts not apparent in tokens or parts-of-speech. Our proposoal is that examining a time-series' second-order moments, which better quantifies changes in linguistic data, enables the investigation of deeper shifts in language. These shifts, like the grammatical relations explored here, show how language is put to use in explanation as well as in general communication.

## Acknowledgments

Thanks to Sam Glucksberg for comments and advice on this research. This work was supported by Enterprise Ireland grant #CC-2011-2601-B for the GRCTC project and a Trinity College research studentship to the first author.

## References

Ahmad, K. (2000). Neologisms, nonces and word formation. In Heid, U., Evert, S., Lehmann, E., and Rohrer, C., editors, *The 9th EURALEX International Congress, Volume II*, pages 711–730, Munich: Universitat Stuttgart.

Ahmad, K. (2011). The "return" and "volatility" of sentiments: An attempt to quantify the behaviour of the markets? In Ahmad, K., editor, *Affective Computing and Sentiment Analysis*. Springer.

Davies, M. (2010). The corpus of contemporary american english as the first reliable monitor corpus of english. *Literary and Linguistic Computing*, 25(4):447–464.

Geeraerts, D. (2002). The scope of diachronic onomasiology. In Agel, V., Gardt, A., Hass-Zumkehr, U., and Roelcke, T., editors, *Das Wort. Seine strukturelle und kulturelle Dimension. Festschrift für Oskar Reichmann zum 65*, pages 29–44. Geburtstag.

Gerow, A. and Keane, M. T. (2011). Mining the web for the "voice of the herd" to track stock market bubbles. In *Proceedings of the 22nd International Joint Conference on A.I.*, Barcelona, Spain.

Griffiths, T. and Steyvers, M. (2004). Finding scientific topics. *Proceedings of the National Academy of Sciences*, 101(1):5228–5235.

Halliday, M. A. K. and Martin, J. R. (1993). *Writing Science: Literacy and Discursive Power*. The Falmer Press, London and Washington DC.

Havre, S., Hetzler, E., Whitney, P., and Nowell, L. (2002). Themeriver: Visualizing thematic changes in large document collections. *IEEE Transactions on Visualization and Computer Graphics*, 8(1):9–20.

Kilgarriff, A., Rychlý, P., Smrž, P., and Tugwell, D. (2004). The sketch engine. In *Proceedings of EURALEX 2004*, pages 105–116.

Kim, D. I. and Sudderth, E. B. (2011). The doubly correlated nonparametric topic model. *Neural Information Processing Systems*, 24.

Kouloumpis, E., Wilson, T., and Moore, J. (2011). Twitter sentiment analysis: The good the bad and the omg! In *Proceedings of the Fifth International AAAI Conference on Weblogs and Social Media*, pages 538–541.

Mair, C., Hundt, M., Leech, G., and Smith, N. (2003). Short term diachronic shifts in part-of-speech frequencies: a comparison of the tagged lob and f-lob corpora. *International Journal of Corpus Linguistics*, 7(2):245–264.

McMahon, A. S. (1994). *Understanding Language Change*. Cambridge University Press, Cambridge and New York.

Michel, J.-B., Shen, Y. K., Aiden, A. P., Veres, A., Gray, M. K., Team, T. G. B., Pickett, J. P., Hoiberg, D., Clancy, D., Norvig, P., Orwant, J., Pinker, S., Nowak, M. A., and Aiden, E. L. (2011). Quantitative analysis of culture using millions of digitized books. *Science*, 331(6014):176–182.

Rosen-Zvi, M., Chemudugunta, C., Griffiths, T., Smyth, P., and Steyvers, M. (2010). Learning author-topic models from text corpora. *ACM Transactions on Information Systems*, 28(1).

Schmid, H. (1994). Probabilistic part-of-speech tagging using decision trees. In *Proceedings of International Conference on New Methods in Language Processing*, Manchester, UK.

Taylor, S. J. (2005). *Asset Price Dynamics, Volatility, and Prediction*. Princeton University Press, Cambridge, MA.

Tetlock, P. (2007). Giving content to investor sentiment: The role of media in the stock market. *The Journal of Finance*, 62(3):1139–1168.